\documentclass[letterpaper]{article} 
\usepackage{aaai25}  
\usepackage{times}  
\usepackage{helvet}  
\usepackage{courier}  
\usepackage[hyphens]{url}  
\usepackage{graphicx} 
\usepackage{natbib}  
\usepackage{caption} 
\usepackage{soul}
\usepackage{url}
\usepackage{amsmath,amsfonts,amssymb}
\usepackage{amsthm}
\usepackage{booktabs}
\usepackage{algorithm}
\usepackage{algpseudocode}
\usepackage[switch]{lineno}
\usepackage{xcolor}
\usepackage{color, colortbl}
\usepackage{amsfonts}
\usepackage{subcaption}
\usepackage{graphicx}
\usepackage{multirow}
\newtheorem{theorem}{\textbf{Theorem}}

\newtheorem{assumption}{\textbf{Assumption}}

\usepackage{newfloat}
\usepackage{listings}
\DeclareCaptionStyle{ruled}{labelfont=normalfont,labelsep=colon,strut=off} 
\floatstyle{ruled}
\newfloat{listing}{tb}{lst}{}
\floatname{listing}{Listing}
\title{WHALE-FL: Wireless and Heterogeneity Aware Latency Efficient Federated Learning over Mobile Devices via Adaptive Subnetwork Scheduling}
\author {
    Huai-an Su\textsuperscript{\rm 1\rm *},
    Jiaxiang Geng\textsuperscript{\rm 2\rm \thanks{These authors contributed equally.}},
    Liang Li\textsuperscript{\rm 3},
    Xiaoqi Qin\textsuperscript{\rm 2}, \\
    Yanzhao Hou\textsuperscript{\rm 2},
    Hao Wang\textsuperscript{\rm 4},
    Xin Fu\textsuperscript{\rm 1},
    Miao Pan\textsuperscript{\rm 1}
}

\affiliations {
    \textsuperscript{\rm 1}Department of Electrical and Computer Engineering, University of Houston\\
    \textsuperscript{\rm 2}School of Information and Communication Engineering, Beijing University of Posts and Telecommunications\\
    \textsuperscript{\rm 3}Frontier Research Center, Peng Cheng Laboratory\\
    \textsuperscript{\rm 4}	Department of Electrical and Computer Engineering, Stevens Institute of Technology
    
    \{hsu3, xfu8, mpan2\}@uh.edu, 
    \{lelegjx, xiaoqiqin, houyanzhao\}@bupt.edu.cn, lil03@pcl.ac.cn,
    hwang9@stevens.edu
}

\begin{document}

\maketitle

\begin{abstract}
As a popular distributed learning paradigm, federated learning (FL) over mobile devices fosters numerous applications, while their practical deployment is hindered by participating devices' computing and communication heterogeneity. Some pioneering research efforts proposed to extract subnetworks from the global model, and assign as large a subnetwork as possible to the device for local training based on its full computing capacity. Although such fixed size subnetwork assignment enables FL training over heterogeneous mobile devices, it is unaware of (i) the dynamic changes of devices' communication and computing conditions and (ii) FL training progress and its dynamic requirements of local training contributions, both of which may cause very long FL training delay. Motivated by those dynamics, in this paper, we develop a \underline{w}ireless and \underline{h}eterogeneity \underline{a}ware \underline{l}atency \underline{e}fficient FL (WHALE-FL) approach to accelerate FL training through adaptive subnetwork scheduling. Instead of sticking to the fixed size subnetwork, WHALE-FL introduces a novel subnetwork selection utility function to capture device and FL training dynamics, and guides the mobile device to adaptively select the subnetwork size for local training based on (a) its computing and communication capacity, (b) its dynamic computing and/or communication conditions, and (c) FL training status and its corresponding requirements for local training contributions. Our evaluation shows that, compared with peer designs, WHALE-FL effectively accelerates FL training without sacrificing learning accuracy.
\end{abstract}

%

\section{Introduction}%
Federated Learning (FL)~\cite{mcmahan2016comm} recently experienced a notable evolution, expanding its scope from conventional data center environments to harness the potential of mobile devices~\cite{9488839,10.1145/3581791.3596865}. This shift has been propelled by the continuous advancements in hardware, empowering mobile devices like the NVIDIA Xavier, iPhone 16, etc. with increasingly robust on-device computing capabilities tailored for local training. With the collective intelligence of edge devices and FL's fundamental principle of preserving data privacy, FL over mobile devices has paved the way for a diverse spectrum of innovative mobile applications, including keyboard predictions~\cite{Hard2018FederatedLF}, smart home hazard detection~\cite{9097597}, health event detection~\cite{BRISIMI201859}, and so on.

While FL over mobile device has great potentials, its practical deployment faces significant challenges due to the inherent heterogeneity among real-world mobile devices, varying in computing capability, wireless conditions and local data distribution~\cite{273723}. Existing FL studies often assume the model-homogeneous setting, where global and local models share identical architectures across all clients. However, as devices are forced to train models within their individual capability, developers have to choose between excluding low-tier devices, introducing training bias~\cite{doi:10.1126/science.187.4175.398}, or maintaining a low-complexity global model to accommodate all clients, resulting in degraded accuracy~\cite{cho2021client,8964354}. The trend towards large models like Transformers~\cite{Liu2023FedETAC} exacerbates the issue, hindering their training on mobile devices. Furthermore, unlike GPU clusters with stable high-speed Internet connections, mobile devices' computing resources are constrained and heterogeneous, and their wireless transmissions are relatively slow and dynamic, both of which lead to huge latency in FL training~\cite{Chen2022FedOBDOB} and may severely degrade the performance of associated applications.

To address the limitations of model-homogeneous FL, researchers have recently studied how to train different sized models across heterogeneous mobile clients and corresponding global model aggregation in FL training.
Subnetwork training, exemplified by pioneering approaches like width-based subnetwork generation in Federated Dropout \cite{Wen2021FederatedDS} and HeteroFL \cite{diao2021heterofl}, and depth-based generation in DepthFL \cite{kim2023depthfl}, has proven effective by enabling mobile devices to train smaller subnetworks derived from the large global server model.
These designs also offer solutions to aggregating diverse devices' subnetworks. By tailoring subnetwork architecture for the individual device, subnetwork training can ensure compatibility with mobile devices owning heterogeneous computing and communication capabilities. However, a prevalent challenge in current subnetwork approaches lies in their static fixed-size subnetwork assignment policy. Such a policy may fail to unleash the full potential of subnetwork based training, mainly due to the unawareness of system dynamics (i.e., computing and communications dynamics) and FL training dynamics.

\textit{System dynamics} encompass the time-varying computing loads of devices' background applications and the fluctuating wireless communication conditions across FL training rounds, which affects the sizes of subnetworks that a mobile device can support over rounds. Since most modern mobile devices (e.g., smartphones) participating in FL training have the ability to run multiple tasks simultaneously~\cite{BANABILAH2022103061}, the dynamic orchestration of CPU/GPU resources across these concurrent activities results in the fluctuations in computing power and available memory for FL tasks, consequently impacting the supported subnetwork sizes for on-device computing. Similarly, wireless communications dynamics caused by users' mobility, wireless channel fading, etc. lead to dynamic transmission rates, which directly affect candidate subnework sizes that a mobile device can support for local model updates.

\textit{FL training dynamics} represents FL convergence's dynamic requirements for the contributions from local training at different training stages, which implicitly affects participating devices' selections on subnetwork sizes. 
Recent studies have revealed that critical learning periods (CLP) exist in the training process of deep neural networks~\cite{Achille2018CriticalLP,Yan_Wang_Li_2022}. As the FL training proceeds, the training contributions from each client gradually decrease. Thus, it is crucial to assign suitable subnetwork sizes to clients based on FL training dynamics.

We observe that failing to capture system or training dynamics and always using the possible largest-sized subnetworks under devices' full capabilities may significantly prolong the FL training process. Different from prior static fixed-size subnetwork assignment methods, in this paper, we propose a \underline{w}ireless and \underline{h}eterogeneity \underline{a}ware \underline{l}atency \underline{e}fficient FL (WHALE-FL) approach to accelerate FL training via adaptive width-wise subnetwork scheduling. WHALE-FL characterizes system dynamics and FL training dynamics and tailors appropriate-sized subnetworks for heterogeneous mobile devices under dynamic computing/wireless environments at different FL training stages. As far as we know, WHALE-FL is the first paper that converts static fixed-size subnetwork allocation, e.g., HeteroFL~\cite{diao2021heterofl}, Federated Dropout \cite{Wen2021FederatedDS}, etc., into adaptive subnetwork scheduling for each device by jointly considering system heterogeneity and FL training dynamics, and conducts system-level experiments for validation. Our salient contributions are summarized as follows:
\begin{itemize}
    \item We design a novel subnetwork selection utility function to capture system and FL training dynamics, guiding mobile devices to adaptively size their subnetworks for local training based on the time-varying computing/communication capacity and FL training status.
    \item We develop a WHALE-FL prototype and evaluate its performance with extensive experiments. The experimental results validate that WHALE-FL can remarkably reduce the latency for FL training over heterogeneous mobile devices without sacrificing learning accuracy.
\end{itemize}

\section{Preliminary}
\subsection{FL over Heterogeneous Mobile Devices}

Consider that $M$ mobile devices in a wireless network collaboratively engage in FL to train a deep neural network on locally distributed datasets $\{L_1,\cdots,L_i,\cdots,L_M\}$. Their local models are parameterized by $\{W_1,\cdots,W_i,\cdots,W_M\}$, which are updated using stochastic gradient descents~\cite{Ruder2016AnOO} on the local data samples through local training. The server collects the local model updates and aggregates them into a global model $W_g$ using model averaging~\cite{mcmahan2016comm,Li2020On}. This aggregation occurs over multiple communication rounds, with the global model at the $r$-th round denoted as $W_g^r = \frac{1}{M}\sum_{m=1}^M W_m^r$. In the subsequent training round, $W_g^r$ is transmitted to mobile devices, and their local models are updated as $W_i^{r+1} = W_g^r$. This process repeats until FL converges, while system heterogeneity (communications and computing) among mobile devices incurs huge training latency and significantly slows down FL convergence.

\subsection{FL with Subnetwork Extraction}\label{sec:heterFLsubnetwork}

To address the system heterogeneity issue in FL training, the subnetwork training method was introduced in~\cite{diao2021heterofl}, which extracts subnetworks in different sizes from the global model. 

Let $\mathcal{W}^P = \{W^1,W^2,\cdots,W^p,\cdots,W^P\}$ be a collection of candidate subnetworks to be selected by mobile devices for local training, where $P$ complexity/size levels are considered. A lower size level $p$ corresponds to a larger-sized subnetwork, and $W^P$ is the smallest subnetwork for selection, i.e., $W^P\subset W^{P-1}\subset \cdots \subset W^1$. We follow the same approach as illustrated in~\cite{diao2021heterofl} to extract subnetworks from the global model by shrinking the width of hidden channel with specific ratios. Let $s \in (0,1]$ be the hidden channel shrinkage ratio. Then, we have $|W^p|/|W_g| = |W^p|/|W^1| = s^{2(p-1)}$. 
With this construction, different sized subnetworks can be assigned to participating mobile devices according to their corresponding capabilities. Suppose that the number of devices in each subnetwork size level is $\{M_1,\cdots,M_P\}$. The server has to aggregate the heterogeneous subnetworks in every training round. As demonstrated in~\cite{diao2021heterofl}, the global aggregation is conducted as follows.
{\small
\begin{align}\label{eq:HeteroFLaggre1}
    W_g = W_g^1 &= W_g^P \cup (W_g^{P-1} \backslash W_g^P) \cup \cdots \cup (W_g^1 \backslash W_g^2)  \nonumber\\
    &=W_g^P \cup \bigcup^P_{p=2}W_g^{p-1} \setminus W_g^{p},
\end{align} where
\begin{subequations}  \label{eq:HeteroFLaggre2}
\begin{align}
    W_g^P = \frac{1}{M}&\sum\limits_{m=1}^M W_m^P, \nonumber\\
    W_g^{p-1} \backslash W_g^p &= \frac{1}{M-M_{p:P}}\sum\limits_{m=1}^{M-M_{p:P}} W_m^{p-1}\backslash W_m^p, \forall p\in[2,P].\nonumber
\end{align}
\end{subequations} }
In this way, each parameter is averaged from those devices whose assigned subnetwork contains that specific parameter, which enables the global aggregation and FL training with different sizes of subnetworks. Although the subnetwork method in~\cite{diao2021heterofl} alleviates the system heterogeneity issue, it is a fixed policy. It cannot capture the dynamic changes of wireless transmission/on-device computing conditions, or the dynamic requirements of contributions from local training at different FL training stages, either of which may result in a huge training latency.

\subsection{Fisher Information}

\noindent Fisher information is utilized as a measurement of how much a change in weights can affect the output of neural networks~\cite{Achille2018CriticalLP}. Fisher information is a 2nd-order approximation of the Hessian of the loss function~\cite{Amari2000MethodsOI,Martens2014NewIA}, providing information on the curvature of the loss landscape near the current weights. Such characteristics help to indicate how fast the gradient changes during training, which may be used to characterize the training dynamics from local device side and further help clients decide how to adjust their subnetwork sizes.

To enable distributed subnetwork scheduling, we use the Federated Fisher Information Matrix (FedFIM) from~ \cite{Yan_Wang_Li_2022} instead of the traditional definition of the Fisher Information Matrix (FIM) for centralized training to avoid requiring access to the entire dataset. That is, given that training data resides in each client, the local FIM on client $i$ in the $r$-th training round is calculated by
\begin{equation}\label{eq:fishertrace}
    FI_{i,r} = \mathbb{E}_{x_i \sim \mathcal{X}_i}\mathbb{E}_{\hat{y} \sim p_W(\hat{y_i}|x_i)}[\bigtriangledown(x_i, \hat{y}_i)\bigtriangledown(x_i, \hat{y}_i)^{\intercal}],
\end{equation}
where $x_i$ is the input data of and $y_i$ is the corresponding output label of client $i$, $W$ is the weight and $p_W(\hat{y}_i|x_i)$ is the approximate posterior distribution. $\mathcal{X}_i$ is the empirical distribution of the $i$-th client's local data. The corresponding gradient of the loss for $(x, y)$ is denoted as $\bigtriangledown(x,y)= \frac{\partial}{\partial W}\ell(x,y;W)$, and $\hat{y_i}$ is a random variable rather than a true label with its distribution following $p_W(\hat{y}_i|x_i)$.

\section{Motivation}\label{Sec:Motivation}

\textbf{Unawareness of system dynamics.} Traditional subnetwork assignment (e.g., HeteroFL~\cite{diao2021heterofl}) is fixed, which is based on the participating mobile device's maximum system capability (i.e., computing + communications), while ignoring the dynamic changes of the device's computing and communication conditions. Such an unawareness may lead to poor subnetwork assignment decisions and significantly delay the FL training process. For instance, a mobile device capable of computing a full-sized model may be experiencing poor wireless access (e.g., 4G/LTE) or running some computing-intensive background applications (e.g., GPU-intensive gaming) in a certain training round. In this case, the fixed full-sized subnetwork assignment may make this device a straggler and cause a big latency in FL training. Thus, \textit{an adaptive subnetwork scheduling aware of system (computing + communication) dynamics is in need.}

\noindent\textbf{Unawareness of FL training dynamics.} The fixed subnetwork assignment is unaware of FL training progress and its dynamic requirements of learning contributions from local mobile devices. 
Since FL training starts from scratch, any contributions from any device's local training will be helpful. Using small-sized subnetworks can expedite on-device computing and wireless transmissions of local model updates. As FL training proceeds into the CLP, more accurate local model updates are needed for the global training model to converge. When FL training is close to convergence (i.e., the late stage), most mobile devices have already made substantial contributions to the global model. For those devices, sticking to large or full-sized subnetworks for local training offers limited learning benefits for FL convergence, while some computing/communications-constrained devices may incur significant training latency or even become stragglers. Therefore, it is necessary to develop an adaptive subnetwork scheduling method that captures FL training dynamics, recognizes computing/communication constraints, and selects appropriately sized subnetworks for local training, to improve delay efficiency in FL training over mobile devices.

\section{WHALE-FL Design}
Aiming to reduce FL training latency, WHALE-FL entitles mobile devices to distributedly schedule different sizes of subnetworks for local training, adapting to their system dynamics and FL training dynamics. To capture those dynamics, WHALE-FL presents a novel adaptive subnetwork selection utility function jointly considering system efficiency and FL training efficiency. Moreover, WHALE-FL provides a normalization procedure to convert the calculated subnetwork selection utility values to discrete size levels of subnetowrks for mobile devices' local scheduling decisions.

\subsection{Adaptive Subnetwork Selection Utility}
WHALE-FL's adaptive subnetwork selection performance hinges on two critical aspects: system efficiency and training efficiency. System efficiency encompasses the duration of each training round, including local computing and model uploading time consumption. Training efficiency gauges the local training's contributions to global convergence. The fluctuating wireless conditions and available computing resources of devices, as well as their training progress with local data, collectively determine the system and training efficiency, forming what we term as \textit{adaptive subnetwork selection utility}.


To accelerate FL training without sacrificing learning accuracy, it is critical to trade-off system and training efficiencies to select the appropriate subnetwork size for individual device's local training per round. Briefly, WHALE-FL favors system efficiency over training efficiency at the early stage of FL training, and tends to schedule small-sized subnetworks for devices' local training. While FL training steps into the middle stage, if more accurate local training is needed for FL convergence, WHALE-FL prefers training efficiency to system efficiency and schedules to adaptively increase the size of subnetworks for participating mobile devices. Otherwise, WHALE-FL prioritizes system efficiency over training efficiency. When FL is close to convergence, WHALE-FL jointly considers system and training efficiencies, and gradually decreases the size of subnetworks for local training, given the fact that most devices have contributed enough to the global model and it is unnecessary to keep large-sized subnetworks for local training.

\noindent\textbf{System efficiency utility.}
We define the system efficiency ($SE_{i,r}$) for any given client $i$ in the $r$-th round based on its wireless transmission rate and available computing resources at that time, which is calculated as follows:
\begin{equation}\label{ResEffi}
    SE_{i,r} =\frac{T}{T_{i,r}^{tr} + T_{i,r}^{co}}, 
\end{equation} where $T_{i,r}^{tr}$ and $T_{i,r}^{co}$ are the transmission delay and the computing delay, respectively, for the unit/smallest subnetwork. $T$ is the developer-preferred duration of each round, which may vary for different FL tasks.
We assume that the wireless transmission rates and available computing resources dynamically change over rounds, but are relatively stable within a FL training round. Thus, given a learning task, transmission and computing workloads for the unit subnetwork are fixed, and $T_{i,r}^{tr}$ and $T_{i,r}^{co}$ can be easily estimated for device $i$ in the $r$-th round. A higher $SE_{i,r}$ enables devices to opt for larger subnetwork sizes for local training within this round, and vice versa. The formulation in Eqn.~(\ref{ResEffi}) comprehensively covers the system efficiency for communication delay dominant cases (i.e., slow transmissions \& fast computing), computing delay dominant cases (i.e., fast transmissions \& slow computing), and communication-computing comparable cases.

\noindent \textbf{Training efficiency utility.} By employing fisher information $FI$, we define the training efficiency utility $TE_{i,r}$ for device $i$ in the $r$-th round as follows:{ \small
\begin{flalign}\label{TraEffi}
   &TE_{i,r} = |\mathbf{B}_i|\sqrt{\frac{1}{|\mathbf{B}_i|} \sum_{k\in \mathbf{B}_i}\sum\limits_{d\in \mathbf{D}} \frac{FI_{i,r-d}(k)^2}{|\mathbf{D}|}},
\end{flalign} }
where $B_i$ represents the batched datasets for device $i$. Here, we utilize a window-averaged local Fisher information to measure the dynamic utility during training with $\mathbf{D}=\{1,..,d,..,D\}$ as the set of window sizes. Here, the sliding window operation helps to prevent frequent zigzag changes in subnetwork sizes, as Fisher information across different local training iterations within the $i$-th round may be highly unstable, and directly using the Fisher information of each iteration could result in unstable subnetwork selection strategies~\cite{Achille2018CriticalLP}. 

\noindent\textbf{Adaptive subnetwork selection utility function.} WHALE-FL trades-off the system and training efficiencies to determine the utility values for subnetwork scheduling over rounds. The adaptive subnetwork selection utility function is shown in Eqn.~(\ref{eq:SSUtilFunc}), where $Util(i,r)$ associates system and training efficiencies with developer-specified factor $\beta$. Aware of both system and FL training dynamics, a large/small value of $Util(i,r)$ suggests that device $i$ should opt for a large/small sized subnetwork in the subsequent $r$-th round. 

{\small
\begin{flalign}\label{eq:SSUtilFunc}
   &Util(i,r) = \nonumber\\
   &{\underbrace{ |\mathbf{B}_i|\sqrt{\frac{1}{|\mathbf{B}_i|} \sum_{k\in \mathbf{B}_i}\sum\limits_{d \in \mathbf{D}} \frac{FI_{i,r-d}(k)^2}{|\mathbf{D}|}} }_{\textit{Training efficiency utility}} }\, \times \underbrace{\left(\frac{T}{T_{i,r}^{tr}+T_{i,r}^{co}}\right)^\beta}_{\textit{System efficiency utility}}.
\end{flalign} }

\subsection{Utility Value to Subnetwork Size Conversion}\label{subsec:conversion}
The calculated utility in Eqn.~(\ref{eq:SSUtilFunc}) cannot directly be used by individual mobile devices to decide their subnetwork size selection. To facilitate mobile devices' decisions, it is necessary to convert subnetwork selection utility values into available/candidate subnetwork sizes.

Given the definitions above, the next step is to normalize devices' utility values into the range of $[0,1]$, in order to identify the model shrinkage ratio. We propose to use a piecewise linear function to normalize $Util(i,r)$ into $U_{n}(i,r)$ as follows.  
\begin{equation}\label{eq:linear}
U_{n}(i,r) = 
\begin{cases}
\frac{Util(i,r)}{U_{th}}, & Util(i,r) \leq U_{th},\\
1, & \text{otherwise},
\end{cases}
\end{equation}
where $U_{th}$ is a configurable threshold that represents the utility level at which the full-sized model should be adopted.

After the utility value normalization, device $i$ selects its subnetwork for the $r$-th round local training by
\begin{equation}\label{eq:min_max_W}
W(i,r) = 
\begin{cases}
\Hat{W}(i,r), &\textbf{if } |W^{max}_i|>|\Hat{W}(i,r)|;\\
W^{max}_{i}, &\textbf{if } |W^{max}_i|\leq|\Hat{W}(i,r)|.
\end{cases}
\end{equation}
Here, $|W^{max}_i|$ denotes the maximum subnetwork size that device $i$ can support with its full computing capacity, where $W^{max}_i\in \mathcal{W}^{P}$ as defined in Sec.~\ref{sec:heterFLsubnetwork}. $\hat{W}(i,r)\in \mathcal{W}^{P}$ is a subnetwork derived from normalized utility value $U_{n}(i,r)$, which can be expressed as
\begin{equation}\label{eq:quantization}
\hat{W}(i,r) =
\begin{cases}
W^1, &\textbf{if } U_{n}(i,r) \geq \frac{(P-1)}{P};\\
W^2, &\textbf{if } U_{n}(i,r) \in [\frac{(P-2)}{P}, \frac{(P-1)}{P});\\
\cdots, &\cdots\\
W^p, &\textbf{if } U_{n}(i,r) \in [\frac{(P-p)}{P}, \frac{(P-p+1)}{P});\\
\cdots, &\cdots\\
W^{P}, &\textbf{if } U_{n}(i,r) < \frac{1}{P},
\end{cases}
\end{equation}
where $|W^p|/|W_g| = s^{2(p-1)}, \forall W^p\in \mathcal{W}^P$.

Then, mobile devices conduct local computing according to their selected subnetworks, respectively, followed by transmitting local model updates to the FL server. Following the same aggregation method in~\cite{diao2021heterofl}, the FL server aggregates updated local models with heterogeneous subnetworks and updates the global model as
\begin{equation}\label{eq:aggre}
    W(g,r+1) = W_g^{P,r+1} \cup \bigcup^P_{p=2}W_g^{p-1,r+1} \setminus W_g^{p,r+1}.
\end{equation}

In summary, during FL training, mobile devices collect their local information at runtime, including up-link channel quality, background computational loads, memory usage, training loss, etc. Based on the collected information, at the beginning of the $r$-th training round, each device leverages Eqn.~(\ref{eq:SSUtilFunc}) to trade-off system efficiency and training efficiency, and calculates its adaptive subnetwork selection utility value $Util(i,r)$. The utility value is then normalized into $U_{n}(i,r)$. Device $i$ uses $U_{n}(i,r)$ to determine the subnetwork size and select an appropriate subnetwork for its local training according to Eqn.~(\ref{eq:min_max_W}) and Eqn.~(\ref{eq:quantization}). After that, FL server aggregates locally trained subnetworks with different sizes and updates the global model for the next round training. In the appendix, we provide the convergence analysis for WHALE-FL based on~\cite{wang2023theoretical}, and show that WHALE-FL can converge under adaptive subnetwork size scheduling.

\section{Experimental Setup}\label{Sec:Implementation}
\subsection{WHALE-FL Testbed}
The testbed consists of an FL aggregator and a set of heterogeneous mobile devices as FL clients. A NVIDIA RTX 3090 serves as the FL server, whose memory capacity is $24$ GB. For heterogeneous FL clients, we have incorporated $5$ types of mobile devices, i.e., MacBookPro2018, NVIDIA Jetson Xavier, NVIDIA Jetson TX2, NVDIA Jetson Nano, and Raspberry Pi 4, representing a range of on-device computing capabilities from high to low. The WHALE-FL system involves a total of $20$ mobile devices, $4$ devices per type. Communication between FL clients and the FL server is facilitated through LTE, BlueTooth, and Wi-Fi $5$ transmission environments. The corresponding transmission rates are $80$ Mbps (Wi-Fi $5$), $20$ Mbps (LTE), and $10$ Mbps (BlueTooth $3.0$), respectively. We set hidden channel shrinkage ratio $s=\frac{1}{2}$ and adopt $5$ subnetwork size levels. Accordingly, the model shrinkage ratios for the $5$ size levels (i.e., $p=1,2,\cdots,5$) are $1$, $\frac{1}{4}$, $\frac{1}{16}$, $\frac{1}{64}$, and $\frac{1}{256}$, respectively.

\subsection{Datasets, Models, Parameters and Baselines}
We conduct our experiments with three different FL tasks: image classification, human activity recognition and language modeling. As for the image classification task, we train a CNN on MNIST dataset~\cite{6296535} and a ResNet$18$ on CIFAR$10$ dataset~\cite{He2015DeepRL}. Human activity recognition involves training a CNN on the HAR dataset~\cite{article}, and a Transformer is trained on the WikiText2 dataset~\cite{Devlin2019BERTPO} for the language modeling task. We use the balanced non-IID data partition~\cite{li2021federated}. Take the MNIST dataset as an example, the total number of classes is $10$. Our default setup is that each device has $\sigma = 2$ classes. We apply a similar non-IID setup to other tasks. The Fisher information's window size $|\mathbf{D}|$ = 10. 
We employ the following peer designs for performance evaluation: (i) FedAvg~\cite{mcmahan2016comm}, where all the clients train with full-sized models; (ii) HeteroFL~\cite{diao2021heterofl}, where subnetwork assignments are fixed and aligned with clients' full computation and communication capabilities; (iii) FedDropout~\cite{Wen2021FederatedDS}, which generates subnetworks by choosing the neurons at random; and (iv) FedRolex~\cite{10.5555/3600270.3602422}, which uses a rolling subnetwork extraction scheme in each FL training round. In particular, we compare the peer design with WHALE-FL's corresponding extension, i.e., the integration of the peer design and WHALE-FL, e.g., FedRolex vs WHALERolex.

\begin{figure*}[t!]
     \centering
     \begin{minipage}[b]{0.245\textwidth}
         \centering
         \includegraphics[width=\textwidth]{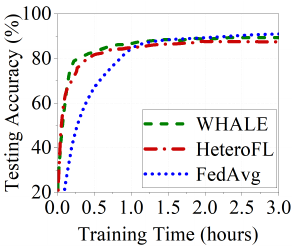}
     \end{minipage}
     \hfill
     \begin{minipage}[b]{0.245\textwidth}
         \centering
         \includegraphics[width=\textwidth]{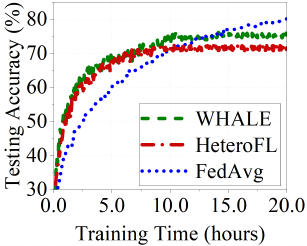}
     \end{minipage}
     \hfill
          \begin{minipage}[b]{0.245\textwidth}
         \centering
         \includegraphics[width=\textwidth]{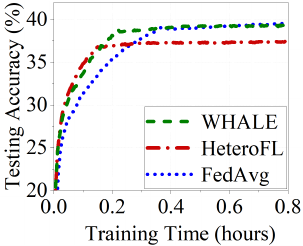}
     \end{minipage}
     \hfill
          \begin{minipage}[b]{0.245\textwidth}
         \centering
         \includegraphics[width=\textwidth]{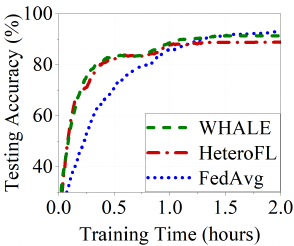}
     \end{minipage}
    \caption{Performance comparison of different FL training approaches under various learning tasks. Figures from left to right are CNN@MNIST, ResNet$18$@CIFAR$10$, Transformer@WikiText$2$, and CNN@HAR with non-IID datasets.}
     \label{task_perf}
\end{figure*}

\begin{table*}[ht!]\centering
\small
\begin{tabular}{c|cccc}
\hline
\multirow{2}{*}{Task} & \multicolumn{2}{c|}{CV}                                              & \multicolumn{1}{c|}{NLP}                   & HAR     \\ \cline{2-5} 
                      & \multicolumn{1}{c|}{CNN@MNIST} & \multicolumn{1}{c|}{Resnet@CIFAR10} & \multicolumn{1}{c|}{Transformer@Wikitext2} & CNN@HAR \\ \hline
Target Accuracy       & \multicolumn{1}{c|}{$85\%$}     & \multicolumn{1}{c|}{$70\%$}          & \multicolumn{1}{c|}{$37\%$}                 & {$88\%$}   \\ \hline
Method                & \multicolumn{4}{c}{Speedup}                                                                                                 \\ \hline
WHALE vs HeteroFL        & \multicolumn{1}{c|}{1.74x}                          & \multicolumn{1}{c|}{1.25x}                               & \multicolumn{1}{c|}{1.21x}                                      & 1.06x   \\ \hline
WHALERolex vs FedRolex        & \multicolumn{1}{c|}{1.75x}                          & \multicolumn{1}{c|}{1.32x}                               & \multicolumn{1}{c|}{1.24x}                                      & 1.10x   \\ \hline
WHALEDropout vs FedDropout      & \multicolumn{1}{c|}{1.70x}                          & \multicolumn{1}{c|}{1.24x}                               & \multicolumn{1}{c|}{1.20x}                                      & 1.05x   \\ \hline
\end{tabular}
\caption{Performance comparison under different subnetwork methods (Speedup).}

\end{table*}

\begin{table*}[ht!]\centering
\small
\begin{tabular}{c|cccc}
\hline
\multirow{2}{*}{Task} & \multicolumn{2}{c|}{CV}                                              & \multicolumn{1}{c|}{NLP}                   & HAR                           \\ \cline{2-5} 
                      & \multicolumn{1}{c|}{CNN@MNIST} & \multicolumn{1}{c|}{Resnet@CIFAR10} & \multicolumn{1}{c|}{Transformer@Wikitext2} & CNN@HAR                       \\ \hline
Method                & \multicolumn{4}{c}{Final Accuracy Improvement}                                                                                                    \\ \hline
FedAvg                & \multicolumn{1}{c|}{92.71\%}                        & \multicolumn{1}{c|}{80.61\%}                             & \multicolumn{1}{c|}{40.54\%}                                    & \multicolumn{1}{c}{92.94\%}                       \\ \hline
HeteroFL $\Rightarrow$ WHALE        & \multicolumn{1}{c|}{87.42\% $\Rightarrow$ 89.29\%}  & \multicolumn{1}{c|}{71.65\% $\Rightarrow$ 75.32\%}       & \multicolumn{1}{c|}{37.40\% $\Rightarrow$ 39.28\%}              & 88.86\% $\Rightarrow$ 91.38\% \\ \hline
FedRolex $\Rightarrow$ WHALERolex        & \multicolumn{1}{c|}{87.82\% $\Rightarrow$ 89.87\%}  & \multicolumn{1}{c|}{72.52\% $\Rightarrow$ 79.57\%}       & \multicolumn{1}{c|}{38.02\% $\Rightarrow$ 39.66\%}              & 89.11\% $\Rightarrow$ 92.03\% \\ \hline
FedDropout $\Rightarrow$ WHALERolex     & \multicolumn{1}{c|}{86.16\% $\Rightarrow$ 87.52\%}  & \multicolumn{1}{c|}{70.08\% $\Rightarrow$ 73.45\%}       & \multicolumn{1}{c|}{37.19\% $\Rightarrow$ 39.06\%}              & 88.25\% $\Rightarrow$ 89.55\% \\ \hline
\end{tabular}
\caption{Performance comparison under different subnetwork methods (Final Accuracy Improvement).}
\label{table:taskperf}
\end{table*}


\section{Evaluation and Analysis}\label{Sec:Result}
\subsection{Latency Efficiency and Learning Performance}
As the results shown in Fig.~\ref{task_perf}, the proposed WHALE-FL consistently achieves remarkable training speedup across various FL tasks without sacrificing learning accuracy. Compared with FedAvg, WHALE-FL accelerates the FL training to the target testing accuracy by approximately 1.5x, 1.9x, 1.3x, and 2.1x for FL tasks including CNN@MNIST, ResNet$18$@CIFAR$10$, Transformer@WikiText2, and CNN@HAR, respectively. As detailed in Sec.~\ref{Sec:Motivation}, HeteroFL's static fixed-size subnetwork assignment policy is not aware of system and training dynamics, which may slow down FL convergence. In contrast, considering both system efficiency and training efficiency, WHALE-FL appropriately assesses the subnetwork selection utility for individual devices and adaptively adjusts the local subnetwork size to suit for time-varying communication and computational conditions and dynamic changing requirements of FL training at different FL training stages, in order to reduce training latency.
Consequently, compared with HeteroFL, WHALE-FL achieves a notable speedup of 1.74x, 1.25x, 1.21x and 1.06x for the tested 4 learning tasks, respectively.
Results in Tables 1 and 2 further demonstrate that WHALE-FL and WHALE-FL based extensions (i.e., WHALEDropout and WHALERolex) achieve faster convergence and better testing accuracy than the peer designs across different FL tasks. 

\subsection{Subnetwork Size and Fisher Information Changes}
\begin{figure*}[ht!]
     \centering
     \begin{minipage}[b]{0.245\textwidth}
         \centering
         \includegraphics[width=\textwidth]{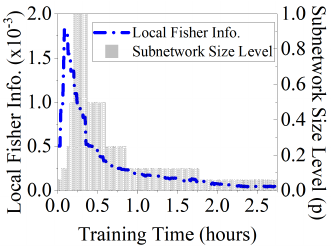}
     \end{minipage}
     \hfill
     \begin{minipage}[b]{0.245\textwidth}
         \centering
         \includegraphics[width=\textwidth]{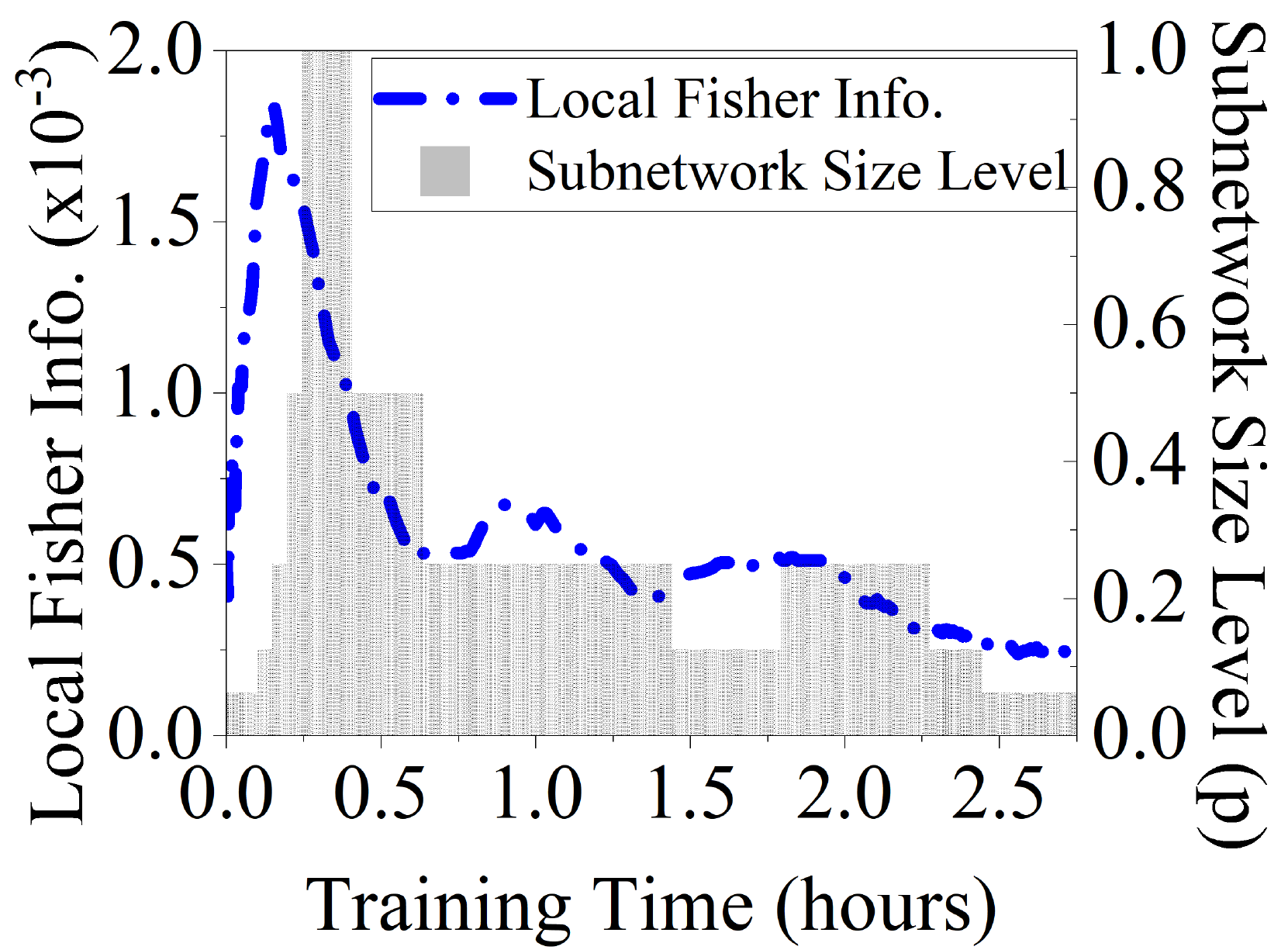}
     \end{minipage}
     \hfill
          \begin{minipage}[b]{0.245\textwidth}
         \centering
         \includegraphics[width=\textwidth]{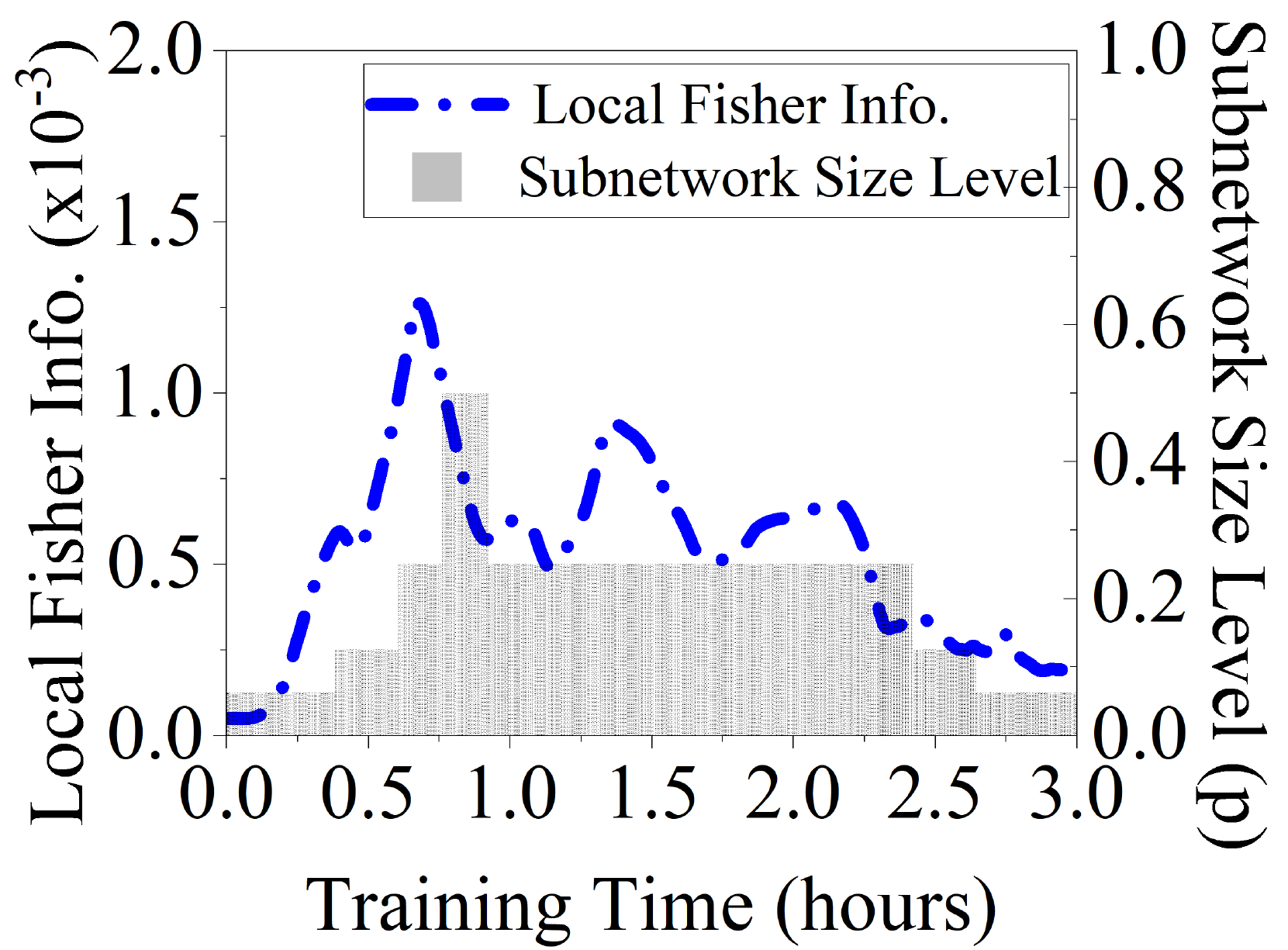}
     \end{minipage}
     \hfill
          \begin{minipage}[b]{0.245\textwidth}
         \centering
         \includegraphics[width=\textwidth]{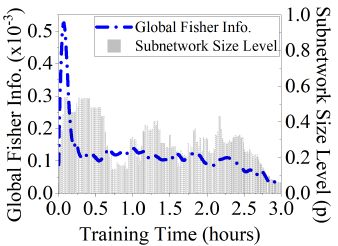}
     \end{minipage}
    \caption{Fisher information and subnetwork size level changes over training time (CNN@MNIST). From left to right, the performance of the user-side models on MacBookPro 2018, NVIDIA Jetson TX2, and Raspberry Pi 4, as well as the global model's performance, are shown.}
     \label{fisher}
\end{figure*}

As shown in Fig.~\ref{fisher}, across the three heterogeneous devices - MacBookPro 2018 (high-end), NVIDIA Jetson TX2 (medium), and Raspberry Pi 4 (low-end) - the subnetwork sizes adjust following the $|\mathbf{D}|$-averaged changes of local Fisher information. The results align with our expectations: When Fisher information is high, the subnetwork size increases to enhance the global model’s accuracy; as training proceeds and Fisher information decreases, indicating that its impacts on learning decrease, the subnetwork is becoming smaller to improve the training time efficiency. On the server side, the averaged size of the aggregated local subnetworks changes along with the global model's Fisher information, which exhibits a similar trend to the local Fisher information. Figure~\ref{fisher} demonstrates that WHALE-FL effectively captures training dynamics while selecting appropriate subnetwork sizes for heterogeneous devices.

\begin{figure}
\centering
\includegraphics[width=0.33\textwidth]{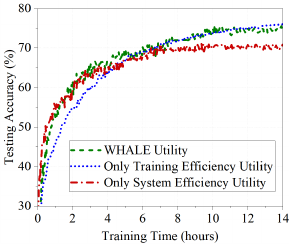}
\caption{Performance comparison of WHALE-FL, system efficiency only and training efficiency only designs (ResNet$18$@CIFAR$10$).}
\label{dynamic_utility_aba}
\end{figure}
\subsection{System Efficiency vs Training Efficiency}

To differentiate system efficiency's contributions from training efficiency's ones, we compare WHALE-FL with system efficiency utility only and training efficiency utility only schedulings. As the results shown in Fig.~\ref{dynamic_utility_aba}, WHALE-FL converges faster than training efficiency only subnetwork scheduling when achieving the target accuracy, since training efficiency only design has no consideration of system dynamics and its impacts on subnetwork size selection; WHALE-FL has better testing accuracy but proceeds slower than system efficiency only subnetwork scheduling at the early training stage. The reason behind is that the system efficiency only design prioritizes system dynamics while ignoring dynamic model accuracy requirements for local training at different FL training stages. WHALE-FL trades-off system and training efficiencies and jointly considers their benefits for FL training.

\begin{figure}[ht!]
     \centering
     \begin{minipage}[b]{0.23\textwidth}
         \centering
         \includegraphics[width=\textwidth]{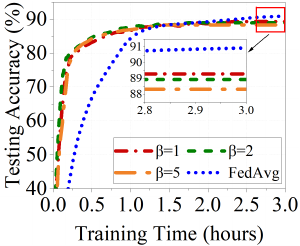}
         \caption*{(a) $\beta$, CNN@MNIST.}
         \label{alpha}
     \end{minipage}
     \hfill
     \begin{minipage}[b]{0.23\textwidth}
         \centering
         \includegraphics[width=\textwidth]{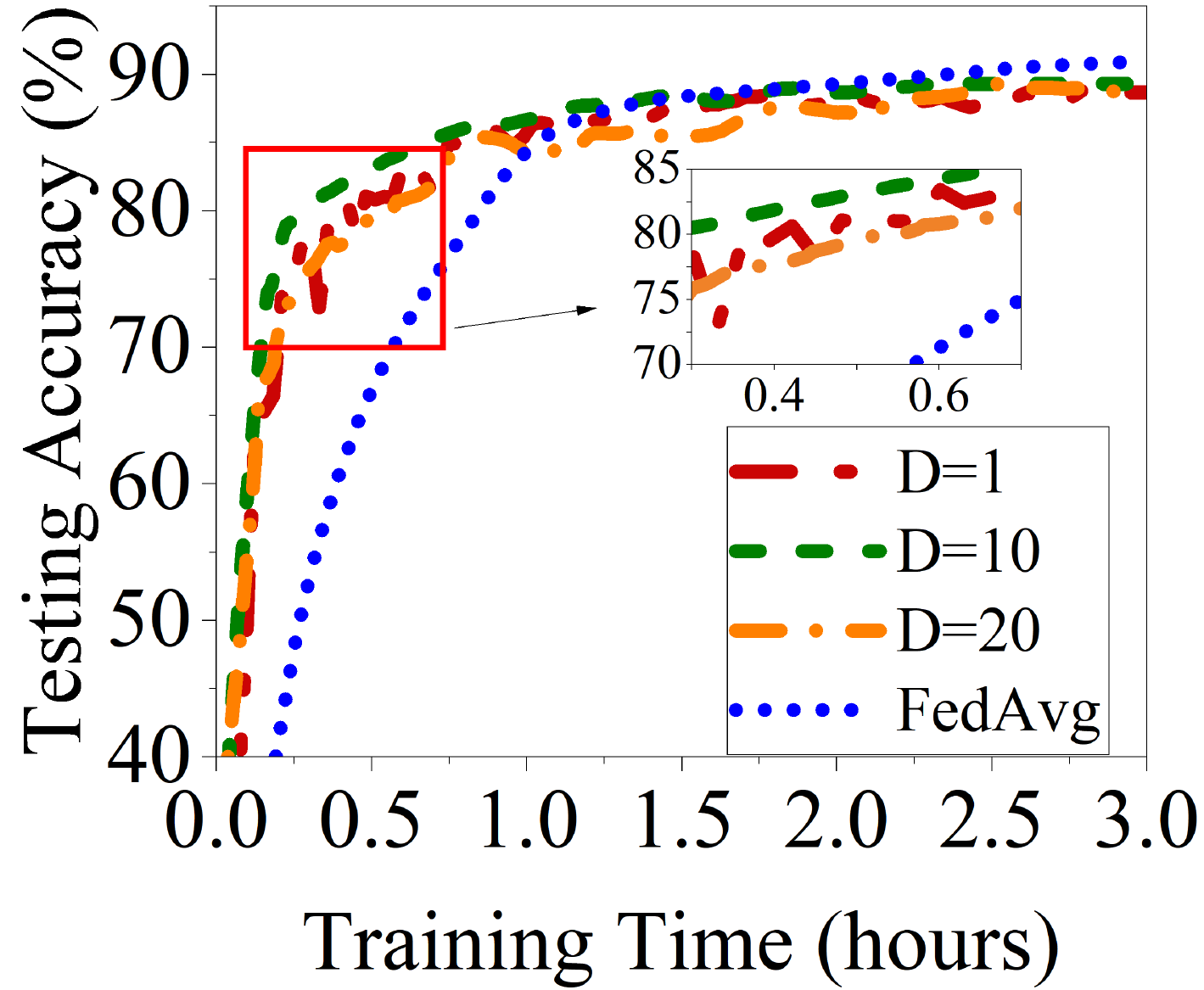}
         \caption*{(b) $D$, CNN@MNIST.}
         \label{subfig:DCNN}
     \end{minipage}
     \hfill
     \begin{minipage}[b]{0.22\textwidth}
         \centering
         \includegraphics[width=\textwidth]{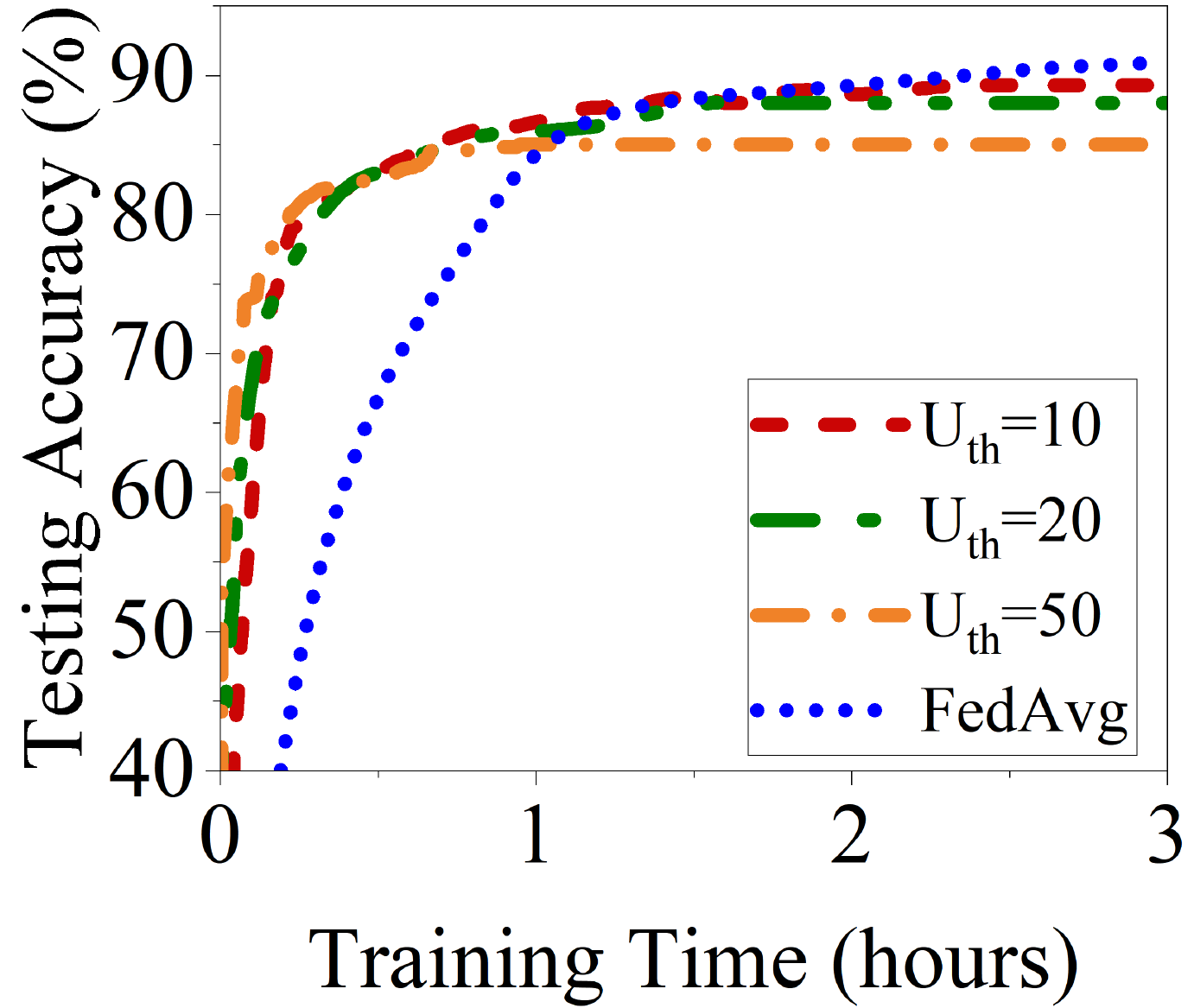}
         \caption*{(c) $U_{th}$, CNN@MNIST.}
     \vspace{+2mm}
     \end{minipage}
     \begin{minipage}[b]{0.23\textwidth}
         \centering
         \includegraphics[width=\textwidth]{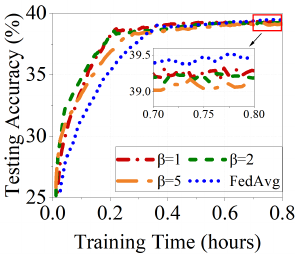}
         \caption*{(d) $\beta$, Transf.@WikiText2.}
     \vspace{+2mm}
     \end{minipage}
       
     \hfill
     \begin{minipage}[b]{0.23\textwidth}
         \centering
         \includegraphics[width=\textwidth]{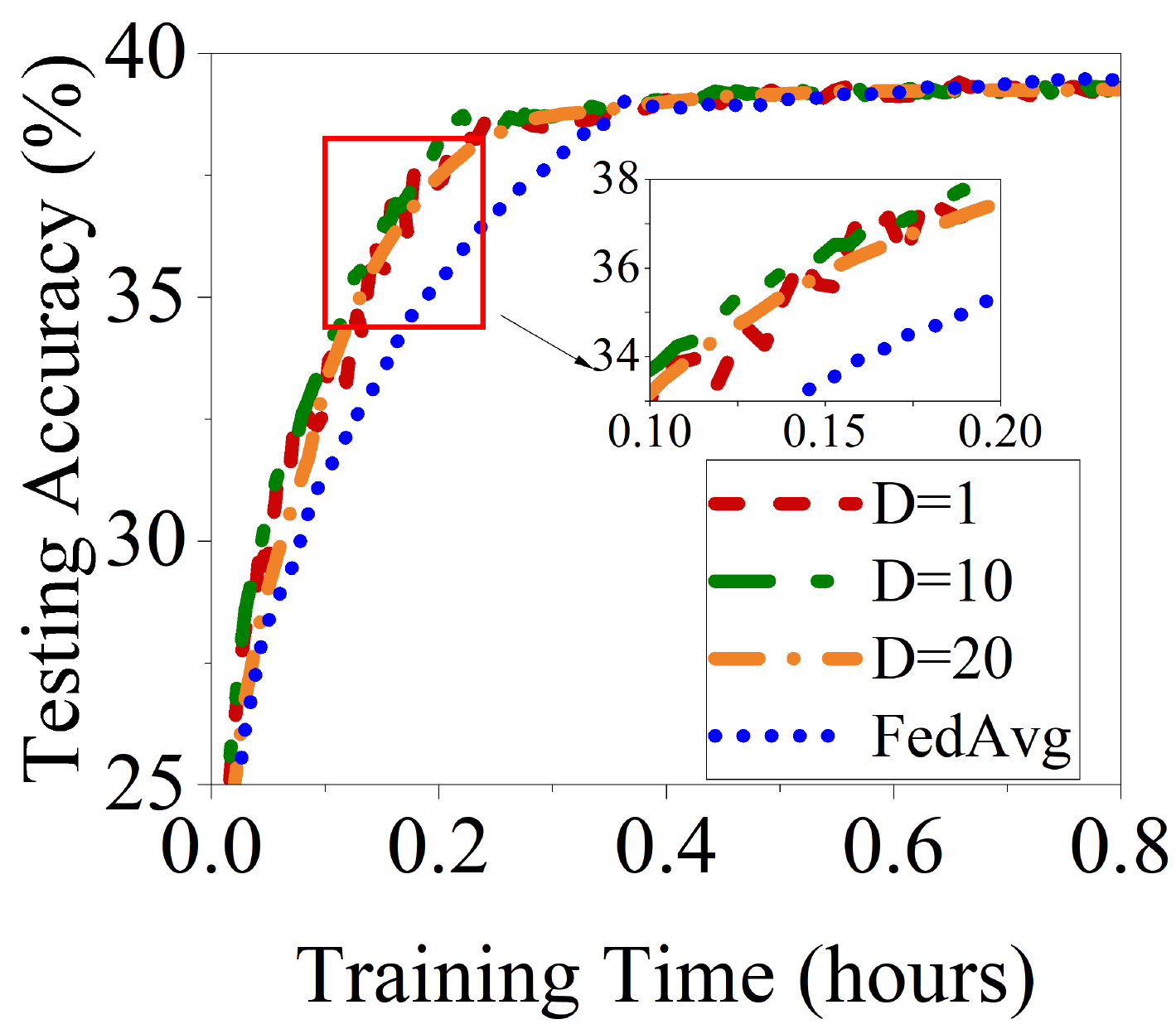}
         \caption*{(e) $D$, Transf.@WikiText2.}
         \label{subfig:DTransf}
     \end{minipage}
     \hfill
     \begin{minipage}[b]{0.23\textwidth}
         \centering
         \includegraphics[width=\textwidth]{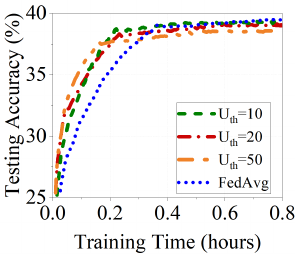}
         \caption*{(f) $U_{th}$, Transf.@WikiText2.}
     \end{minipage}
        \caption{Sensitivity analysis under different $\beta$, $D$, and $U_{th}$ values (a-c: CNN@MNIST; d-e: Transformer@WikiText2).}
     \label{abg}
\end{figure}

\subsection{Sensitivity Analysis}
We further evaluate the impacts of $\beta$, $D$ = $|\mathbf{D}|$, and $U_{th}$, defined in the subnetwork selection utility function, on subnetwork scheduling. 

The hyperparameter $\beta$ trades-off system efficiency and training efficiency utilities. The large/small $\beta$ value means that the device prioritizes system/training efficiency. As the results shown in Fig.~\ref{abg}(a) and Fig.~\ref{abg}(d), we find that the FL training converges slower but achieves higher testing accuracy when $\beta$ is small, e.g., $\beta = 1$, while FL training is faster at early stages but achieves lower testing accuracy when $\beta$ is larger, e.g., $\beta = 5$. System efficiency and training efficiency are somehow balanced when $\beta = 2$. Thus, although $\beta$ is a developer-specified factor, a proper selection of $\beta$ value helps FL training converge fast while achieving good learning performance. 
The hyperparameter $D$ represents the window size for calculating the averaged Fisher information. A small window size, such as $D = 1$ in Fig.~\ref{abg}(b) and Fig.~\ref{abg}(e), makes the subnetwork size updates sensitive to changes in Fisher information, leading to fluctuations in model accuracy during training. 
Conversely, employing a larger window size, like $D = 20$, results in slower subnetwork-size changes. This may cause a situation where a small-sized subnetwork is well-trained while the clients have no chance to switch to the larger-sized subnetworks, thus impairing the training performance during the critical learning periods. A window size of $D = 10$ strikes a good balance, achieving faster convergence.
Similarly, a higher $U_{th}$, e.g., $U_{th}$ = 50 shown in Fig.~\ref{abg}(c) and ~\ref{abg}(f), leads clients to choose smaller subnetworks, which speeds up FL convergence in the early stages by reducing transmission and computation delays but results in lower final accuracy. Conversely, with $U_{th}$ = 10, clients select larger subnetworks, which slows down convergence but yields higher accuracy. A proper $U_{th}$ selection helps to balance learning performance and delay efficiency.

\begin{table}
\small
\centering
\resizebox{\linewidth}{!}{
\begin{tabular}{c|ccc}
\toprule
Local Model     & \multicolumn{3}{c}{CNN@MNIST}                                                      \\ \hline
non-IID Level   & \multicolumn{1}{c|}{$ \sigma=2$} & \multicolumn{1}{c|}{$ \sigma=5$} & $ \sigma=10$ \\ \hline
Target Acc. & \multicolumn{1}{c|}{85\%}        & \multicolumn{1}{c|}{90\%}        & 95\%         \\ \hline
Metric         & \multicolumn{3}{c}{Hours (SP)}               \\ \midrule
FedAvg          & \multicolumn{1}{c|}{1.12 (1.00x)}  & \multicolumn{1}{c|}{0.33 (1.00x)}  & {0.30 (1.00x)}   \\ \hline
HeteroFL           & \multicolumn{1}{c|}{1.06 (1.06x) }  & \multicolumn{1}{c|}{0.26 (1.27x)}  & {0.10 (3.00x)}   \\ \hline
WHALE        & \multicolumn{1}{c|}{\textbf{0.61 (1.84x)}}  & \multicolumn{1}{c|}{\textbf{0.19 (1.74x)}}  & \textbf{{0.08 (3.75x)}}   \\ \hline
FedDropout          & \multicolumn{1}{c|}{1.09 (1.03x)}  & \multicolumn{1}{c|}{0.28 (1.18x)}  &  {0.11 (2.73x)}  \\ \hline
WHALEDropout        & \multicolumn{1}{c|}{\textbf{0.64 (1.75x)}}  & \multicolumn{1}{c|}{\textbf{0.21 (1.57x)}}  &  {\textbf{0.08 (3.75x)}}  \\ \hline
FedRolex           & \multicolumn{1}{c|}{1.03 (1.09x)}  & \multicolumn{1}{c|}{0.25 (1.32x)}  &  {0.10 (3.00x)}  \\ \hline
WHALERolex        & \multicolumn{1}{c|}{\textbf{0.59 (1.90x)}}  & \multicolumn{1}{c|}{\textbf{0.18 (1.83x)}}  &  {\textbf{0.07 (4.29x)}}  \\ 
\bottomrule
\end{tabular}
}
\caption{Performance comparison under different data heterogeneity (CNN@MNIST), where ``SP" is the speedup.}
\label{table:sigma}
\end{table}

\subsection{Impacts of Data Heterogeneity}
We further evaluate the impacts of data heterogeneity on WHALE-FL's performance. Here, we take CNN@MNIST as an example and use the balanced non-IID data partition~\cite{li2021federated}. The total number of classes in the MNIST dataset is $10$. We study the cases where each device has $\sigma = 2, 5$ or $10$ classes, where the data distribution is IID if $\sigma = 10$, i.e., every device has all classes. 
The results are shown in Table~\ref{table:sigma}, where we find that (i) FL training with non-IID data takes longer time to converge, and (ii) embracing both system and training efficiency utilities, WHALE-FL can remarkably improve FL training delay efficiency when applied to existing subnetwork methods under various data heterogeneity scenarios.

\section{Conclusion}
In this paper, we have proposed WHALE-FL, a wireless and heterogeneity aware latency efficient federated learning approach, to accelerate FL training over mobile devices via subnetwork scheduling. Unlike existing static fixed-size subnetwork assignments, WHALE-FL has incorporated an adaptive subnetwork scheduling policy, enabling mobile devices to flexibly select subnetwork sizes for local training, with a keen awareness of mobile devices' system dynamics and FL training dynamics. At its core, WHALE-FL has employed a well-designed subnetwork selection utility function, capturing changes in the device's system conditions (including available computing and communication capacities) and evolving FL training requirements for local training, to schedule appropriate subnetworks for mobile devices in each FL training round. Experimental results have demonstrated that WHALE-FL surpasses peer designs, significantly accelerating FL training over heterogeneous mobile devices without sacrificing learning accuracy.

\section{Acknowledgements}
The work of L. Li was supported in part by the NSFC 62201071. The work of H. Su and M. Pan was supported in part by the US National Science Foundation under grants CNS-2107057, CNS-2318664, CSR-2403249, and CNS-2431596. The work of X. Fu was partially supported by NSF grants CCF-2130688, and CNS-2107057. The work of X. Qin was supported in part by NSFC 62371072 and Beijing Nova Program 2024102. The work of Y. Hou was supported in part by Beijing NSF L232001 and NSFC U21A20449. The work of H. Wang was supported in part by NSF 2315612, 2403247, and 2431595.

\bibliography{aaai25}

\appendix
\onecolumn
\section{Technical Appendix}
\subsection{Convergence Analysis}

Define 
\begin{subequations}  
\begin{align}
S^P &=W_g ^P, \\
S^p &=W_g ^{p-1}\backslash W_g^p, \forall p\in[2,P], \\
\mathcal{S} &=\{S^1,S^2,...,S^p,...,S^P\}
\end{align}
\end{subequations}

Here, $ S^p $ for $ p=1,2,...,P$ represents neural regions such that a subnetwork at any width level $p$ can be constituted by a set of neural regions $\{ S^P, S^{P-1},..., S^p \}$. Let $\mathcal{S}_r$ be the set of the neural regions trained in round $r$, where $\mathcal{S}_r \subset \mathcal{S}$. let $\mathcal{M}_r^i$ be the clients set whose subnetworks train parameters in the neural region $i\in \mathcal{S}_r $ in round $r$ and $|\mathcal{M}_r^i|$ be the number of clients in $\mathcal{M}_r^i$.

\begin{assumption} [Mask-induced noises] Existing $\omega\in[0, 1)$, the mask-induced noise on client $n$ and any $r$ is bounded by:
\begin{equation}
    ||\theta_r - \theta_r \odot m_{r,n}|| \leq \omega_1^2 ||\theta_r||^2
\end{equation}
\end{assumption}

\begin{assumption} [Smoothness Condition] Loss function $ F(\cdot)$ is with $L$-smoothness:
\begin{equation}
    [F(\theta)] - [F(\varphi)] \leq \langle \nabla F(\varphi), \theta - \ varphi \rangle  + \frac{L}{2} \mathbb{E}[||\theta - \varphi ||^2]
\end{equation}
\end{assumption}


\begin{assumption} [Bounded compression]  An operator $\mathcal{C}:\mathbb{R}^d\rightarrow \mathbb{R}^d $ is a $\omega_2$-approximate compressor for $\omega_2\in(0,1]$ if
\begin{equation}
     \mathbb{E} || \mathcal{C}(\theta) - \theta ||^2 \leq \omega_2^2 ||\theta||^2, \quad \forall \theta \in \Omega
\end{equation}
\end{assumption}

\begin{assumption} [Bounded variance] There exists $\sigma>0$, satisfying:
\begin{equation}
     \mathbb{E}_{\xi_{n,t} \sim \mathcal{D}_n} || \nabla F_n(\theta_{r,n,t}; \xi_{n,t}) - \nabla F_n(\theta_{r,n,t}) ||^2 \leq \sigma^2, \quad \forall r, n, t
\end{equation}
\end{assumption}

\begin{assumption} [Bounded data heterogeneity level] There exists $\delta>0$, satisfying:
\begin{equation}
     ||\nabla F_n(\theta_r) - \nabla F(\theta_r)||^2 \leq \delta^2
\end{equation}
\end{assumption}

\begin{theorem} Let all assumptions hold. Suppose that the step size $\gamma$ satisfies $ 0 \leq \gamma \leq \min\left\{ \frac{1}{12TL}, \frac{|\mathcal{M}^*|}{16TL\sqrt{N}}, \left(\frac{|\mathcal{M}^*|}{768T^3L^3N}\right)^{\frac{1}{3}}\right\} $. Then, for all $Q\geq 1$, we have:

\begin{align}\label{ConvRate}
&\frac{1}{R} \sum_{r=1}^R \sum_{i \in S_r} \mathbb{E} ||\nabla F^i(\theta_r)||^2 \nonumber\\
\leq & \frac{8 }{ RT \gamma }\left(\mathbb{E}[F(\theta_1)] - \mathbb{E}[F(\theta_{R+1})] \right) 
+ \left( 64 \omega^2 \frac{N }{|\mathcal{M}^*|}  L^2 + 96 L^3 \gamma T \frac{N }{|\mathcal{M}^*|}  \right) \frac{1}{R}\sum_{r=1}^R \mathbb{E} ||\theta_r||^2 \nonumber\\
&+\frac{8N }{|\mathcal{M}^*|}  \left( 32 \gamma^2 T^2 L^2 + 1 + 96 L^3 \gamma^3 T^3 + 3 L \gamma T \right) \delta^2 + \gamma L \frac{8N }{|\mathcal{M}^*|}  (4 \gamma TL + \frac{3}{2} +12L^2 \gamma^2 T^2) \sigma^2 
\end{align}
\end{theorem} 

Theorem 1 demonstrates the convergence rate of the WHALE-FL algorithm by providing an upper bound on the average gradient of all clients across all trained parameters. WHALE-FL relaxes the constraint that all model parameters must be trained in every round. Equation (\ref{ConvRate}) shows that WHALE-FL can converge under arbitrary adaptive subnetwork size scheduling. The results indicate that larger $|S_r|$ values for each training round $r$ lead to more bounded gradients in the trained neuron regions, improving the convergence rate. Specifically, except for the non-trained parameters from the global model, others can be trained by at least $|\mathcal{M}^*|$ subnetworks in each round. As $|\mathcal{M}^*|$ increases, the model parameters are trained more frequently, allowing WHALE-FL to converge to a stationary point more quickly.

\noindent\textbf{Proof:}

Let's start with the smoothness condition:
\begin{align}
&\mathbb{E}[F(\theta_{r+1})] - \mathbb{E}[F(\theta_r)] \leq \mathbb{E}[\langle \nabla F(\theta_r), \theta_{r+1} - \theta_r \rangle ] + \frac{L}{2} \mathbb{E}[||\theta_{r+1} - \theta_r ||^2].
\end{align}

\vspace{+30mm}
Bound $U_1$:

\begin{flalign}
&\quad \mathbb{E}\langle \nabla F(\theta_r), \theta_{r+1} - \theta_r \rangle \nonumber\\ 
&= \sum_{i \in \mathcal{S}_r} \mathbb{E} \langle \nabla F^i(\theta_r), \theta_{r+1}^i - \theta_r^i \rangle + \sum_{i \in \mathcal{S} \backslash\mathcal{S}_r} \mathbb{E} \langle \nabla F^i(\theta_r), \theta_{r+1}^i - \theta_r^i \rangle \\
&= \sum_{i \in \mathcal{S}_r} \mathbb{E} \langle \nabla F^i(\theta_r), \theta_{r+1}^i - \theta_r^i \rangle + \sum_{i \in \mathcal{S} \backslash\mathcal{S}_r} \mathbb{E} \langle \nabla F^i(\theta_r), 0 \rangle \\
&= \sum_{i \in \mathcal{S}_r} \mathbb{E} \langle \nabla F^i(\theta_r), - \frac{1}{|\mathcal{M}_r^i|} \sum_{n \in \mathcal{M}_r^i} (\theta_{r,n,0} - \theta_{r,n,T}) \rangle \\
&= \sum_{i \in \mathcal{S}_r} \mathbb{E} \langle \nabla F^i(\theta_r), - \frac{1}{|\mathcal{M}_r^i|} \sum_{n \in \mathcal{M}_r^i} (\theta_{r,n,0} - (\theta_{r,n,0}  
-\gamma \sum_{t=1}^{T} \nabla F_n(\theta_{r,n,t-1}, \xi_{n,t-1}) \odot m_{n,r})) \rangle \\
&= \sum_{i \in \mathcal{S}_r} \mathbb{E} \langle \nabla F^i(\theta_r), - \frac{1}{|\mathcal{M}_r^i|} \sum_{n \in \mathcal{M}_r^i} (\gamma \sum_{t=1}^{T} \nabla F_n^i (\theta_{r,n,t-1})) \rangle \\
&= \sum_{i \in \mathcal{S}_r} \mathbb{E} \langle \nabla F^i(\theta_r), - \frac{1}{|\mathcal{M}_r^i|} \sum_{n \in \mathcal{M}_r^i}  (\gamma \sum_{t=1}^{T} \nabla F_n^i (\theta_{r,n,t-1})) \rangle \\
&= \sum_{i \in \mathcal{S}_r} \mathbb{E} \langle \nabla F^i(\theta_r), - \frac{1}{|\mathcal{M}_r^i|} \sum_{n \in \mathcal{M}_r^i} \sum_{t=1}^{T}  \gamma [\nabla F_n^i (\theta_{r,n,t-1}) - \nabla F^i (\theta_r) + \nabla F^i (\theta_r)] \rangle \\
&= - \sum_{i \in \mathcal{S}_r} \gamma T \mathbb{E} || \nabla F^i(\theta_r) ||^2 + \underbrace{\sum_{i \in \mathcal{S}_r} \mathbb{E} \langle \nabla F^i(\theta_r), - \frac{1}{|\mathcal{M}_r^i|} \sum_{n \in \mathcal{M}_r^i} \sum_{t=1}^{T}  \gamma [\nabla F_n^i (\theta_{r,n,t-1}) - \nabla F^i (\theta_r)] \rangle}_{U_3}.
\end{flalign}

\text{Bound } $U_3$:
\begin{flalign}
&\quad \sum_{i \in \mathcal{S}_r} \mathbb{E} \langle \nabla F^i(\theta_r), - \frac{1}{|\mathcal{M}_r^i|} \sum_{n \in \mathcal{M}_r^i}\sum_{t=1}^{T}  \gamma [\nabla F_n^i (\theta_{r,n,t-1}) - \nabla F^i (\theta_r)] \rangle \nonumber\\
& \leq \frac{\gamma T }{2}\sum_{i \in \mathcal{S}_r} \mathbb{E} || \nabla F^i(\theta_r) ||^2 + \frac{\gamma T}{2} \sum_{i \in \mathcal{S}_r} \mathbb{E} || \frac{1}{T|\mathcal{M}_r^i|} \sum_{n \in \mathcal{M}_r^i} \sum_{t=1}^{T} [\nabla F_n^i(\theta_{r,n,t-1}) - \nabla F^i(\theta_r)] ||^2 \\
& \leq \frac{\gamma T }{2}\sum_{i \in \mathcal{S}_r} \mathbb{E} || \nabla F^i(\theta_r) ||^2 +\gamma T \sum_{i \in \mathcal{S}_r} \mathbb{E} || \frac{1}{T|\mathcal{M}_r^i|} \sum_{n \in \mathcal{M}_r^i} \sum_{t=1}^{T} [\nabla F_n^i(\theta_{r,n,t-1}) - \nabla F_n^i(\theta_r)] ||^2 \nonumber\\
&+\gamma T \sum_{i \in \mathcal{S}_r} \mathbb{E} || \frac{1}{T|\mathcal{M}_r^i|} \sum_{n \in \mathcal{M}_r^i} \sum_{t=1}^{T} [\nabla F_n^i(\theta_r) - \nabla F^i(\theta_r)] ||^2\\
& \leq \frac{\gamma T }{2}\sum_{i \in \mathcal{S}_r} \mathbb{E} || \nabla F^i(\theta_r) ||^2 +\gamma T \sum_{i \in \mathcal{S}_r} \mathbb{E} || \frac{1}{T|\mathcal{M}_r^i|} \sum_{n \in \mathcal{M}_r^i} \sum_{t=1}^{T} [\nabla F_n^i(\theta_{r,n,t-1}) - \nabla F_n^i(\theta_r)] ||^2 \nonumber\\
&+ \frac{\gamma }{|\mathcal{M}^*|}   \sum_{n=1}^N \sum_{t=1}^{T} \mathbb{E} || [\nabla F_n (\theta_r) - \nabla F (\theta_r)] ||^2\\
& \leq \frac{\gamma T }{2}\sum_{i \in \mathcal{S}_r} \mathbb{E} || \nabla F^i(\theta_r) ||^2 +\underbrace{\gamma T \sum_{i \in \mathcal{S}_r} \mathbb{E} || \frac{1}{T|\mathcal{M}_r^i|} \sum_{n \in \mathcal{M}_r^i} \sum_{t=1}^{T} [\nabla F_n^i(\theta_{r,n,t-1}) - \nabla F_n^i(\theta_r)] ||^2}_{U_4} + \frac{TN\gamma}{ |\mathcal{M}^*|} \delta^2.
\end{flalign}

\text{Bound } $U_4$:
\begin{flalign}
&\quad \gamma T \sum_{i \in \mathcal{S}_r} \mathbb{E} || \frac{1}{T|\mathcal{M}_r^i|} \sum_{n \in \mathcal{M}_r^i} \sum_{t=1}^{T} [\nabla F_n^i(\theta_{r,n,t-1}) - \nabla F_n^i(\theta_r)] ||^2 \nonumber\\
& \leq \frac{\gamma }{ |\mathcal{M} ^*|} \sum_{n =1}^N \sum_{t=1}^{T}\sum_{i \in \mathcal{S}_r}   \mathbb{E} || \nabla F_n^i(\theta_{r,n,t-1}) - \nabla F_n^i(\theta_r) ||^2\\
& \leq \frac{\gamma }{ |\mathcal{M} ^*|} \sum_{n =1}^N \sum_{t=1}^{T} \mathbb{E} || [\nabla F_n (\theta_{r,n,t-1}) - \nabla F_n (\theta_r)] ||^2\\
& \leq \frac{T \gamma }{ |\mathcal{M} ^*|} \sum_{n =1}^N L^2 \underbrace{\frac{1}{T} \sum_{t=1}^{T} \mathbb{E} || \theta_{r,n,t-1} - \theta_{r,n,0} ||^2}_{U_5}.
\end{flalign}

\text{Bound } $U_5$:
\begin{flalign}
&\quad \frac{1}{T} \sum_{t=1}^{T} \mathbb{E} \left[ || \theta_{r,n,t-1} - \theta_{r,n,0} ||^2 \right] \nonumber\\
& \leq \frac{2}{T} \sum_{t=1}^{T} \mathbb{E}|| \theta_{r,n,t-1} - \theta_{r,n,0} ||^2 + \frac{2}{T} \sum_{t=1}^{T} \mathbb{E} || \theta_{r,n,0} - \theta_{r} ||^2 \\ 
&= \frac{2}{T} \sum_{t=1}^{T} \mathbb{E} || \sum_{j=0}^{t-2} -\gamma \nabla F_n (\theta_{r,n,j}, \xi_{n,j}) \odot m_{r,n} ||^2 + \frac{2}{T} \sum_{t=1}^{T} \mathbb{E} || \mathcal{C}(\theta_r) \odot m_{n,r} - \theta_r ||^2 \\
&= \frac{2 \gamma^2}{T} \sum_{t=1}^{T} \mathbb{E} || \sum_{j=0}^{t-2} (\nabla F_n (\theta_{r,n,j}, \xi_{n,j}) - \nabla F_n (\theta_{r,n,j}) + \nabla F_n (\theta_{r,n,j})) \odot m_{r,n} ||^2 \nonumber\\
&+ \frac{2}{T} \sum_{t=1}^{T} \mathbb{E} || \mathcal{C}(\theta_r) \odot m_{n,r} - \mathcal{C}(\theta_r) + \mathcal{C}(\theta_r) - \theta_r ||^2 \\ 
&\leq \frac{4 \gamma^2}{T} \sum_{t=1}^{T} \mathbb{E} || \sum_{j=0}^{t-2} (\nabla F_n (\theta_{r,n,j}, \xi_{n,j}) - \nabla F_n (\theta_{r,n,j})) \odot m_{r,n} ||^2 + \frac{4 \gamma^2}{T} \sum_{t=1}^{T} \mathbb{E} || \sum_{j=0}^{t-2} \nabla F_n (\theta_{r,n,j}) \odot m_{r,n} ||^2\nonumber\\
& + \frac{4}{T} \sum_{t=1}^{T} \mathbb{E} || \mathcal{C}(\theta_r) \odot m_{n,r} - \mathcal{C}(\theta_r) ||^2  + \frac{4}{T} \sum_{t=1}^{T} \mathbb{E} || \mathcal{C}(\theta_r) - \theta_r ||^2 \\
&\leq \frac{4 \gamma^2}{T} \sum_{t=1}^{T} (t-1)\sigma^2 + \frac{4 \gamma^2}{T} \sum_{t=1}^{T} \mathbb{E} || \sum_{j=0}^{t-2} [\nabla F_n (\theta_{r,n,j}) -\nabla F_n (\theta_{r }) +\nabla F_n (\theta_{r }) \odot m_{r,n} ||^2\nonumber\\
&+ \frac{4}{T} \sum_{t=1}^{T} \omega_1^2 \mathbb{E} || \mathcal{C}(\theta_r) ||^2 + \frac{4}{T} \sum_{t=1}^{T} \omega_2^2 \mathbb{E} || \theta_r ||^2\\
&\leq 2 \gamma^2 T \sigma^2 + \frac{8 \gamma^2}{T} \sum_{t=1}^{T} (t-1) \sum_{j=0}^{t-2} \mathbb{E} || (\nabla F_n (\theta_{r,n,j}) - \nabla F_n (\theta_r)) \odot m_{r,n} ||^2 \nonumber\nonumber\\
&+ \frac{8 \gamma^2}{T} \sum_{t=1}^{T} (t-1) \sum_{j=0}^{t-2} \mathbb{E} || \nabla F_n (\theta_r) \odot m_{q,n} ||^2 + \frac{4}{T} \sum_{t=1}^{T} \omega_1^2 \mathbb{E} || \mathcal{C}(\theta_r) - \theta_r + \theta_q ||^2 + \frac{4}{T} \sum_{t=1}^{T} \omega _2^2 \mathbb{E} || \theta_r ||^2\\
&\leq 2 \gamma^2 T \sigma^2 + \frac{8 \gamma^2 L^2}{T} \sum_{t=1}^{T} (t-1) \sum_{j=0}^{t-2} \mathbb{E} || \theta_{r,n,j} - \theta_r ||^2 + 8 \gamma^2 T^2 \mathbb{E} || (\nabla F_n (\theta_r) - \nabla F (\theta_r) + \nabla F (\theta_r)) \odot m_{r,n} ||^2 \nonumber\\
&+ \frac{8}{T} \sum_{t=1}^{T} \omega_1^2 \mathbb{E} || \mathcal{C}(\theta_r) - \theta_r ||^2 + \frac{8}{T} \sum_{t=1}^{T} \omega_1^2 \mathbb{E} || \theta_r ||^2 + \frac{4}{T} \sum_{t=1}^{T} \omega_2^2 \mathbb{E} || \theta_r ||^2  \\
&\leq 2 \gamma^2 T \sigma^2 + 8 \gamma^2 L^2 T^2 (T-1) \frac{1}{T} \sum_{t=1}^{T} \mathbb{E} || \theta_{r,n,t-1} - \theta_r ||^2 + 16 \gamma^2 T^2 \mathbb{E} || (\nabla F_n (\theta_r) - \nabla F (\theta_r)) \odot m_{r,n} ||^2 \nonumber\\
&+ 16 \gamma^2 T^2 \mathbb{E} || \nabla F (\theta_r) \odot m_{r,n} ||^2 + \frac{8}{T} \sum_{t=1}^{T} \omega_1^2 \omega_2^2 \mathbb{E} || \theta_r ||^2 + \frac{8}{T} \sum_{t=1}^{T} \omega_1^2 \mathbb{E} || \theta_r ||^2 + \frac{4}{T} \sum_{t=1}^{T} \omega_2^2 \mathbb{E} || \theta_r ||^2 \\
&\leq 2 \gamma^2 T \sigma^2 + 8 \gamma^2 L^2 T^2 \frac{1}{T} \sum_{t=1}^{T} \mathbb{E} || \theta_r - \theta_{r,n,t-1} ||^2 + 16 \gamma^2 T^2 \sigma^2 \nonumber\\
&+ 16 \gamma^2 T^2 \mathbb{E} || \nabla F (\theta_r) \odot m_{r,n} ||^2 + 4 (2 \omega_1^2 \omega_2^2 + 2 \omega_1^2 + \omega_2^2) \mathbb{E} || \theta_r ||^2.
\end{flalign}

Let $\omega_1^2 \omega_2^2 + 2 \omega_1^2 + \omega_2^2 = w^2, 8\gamma^2 L^2 T r^2 \leq \frac{1}{2} \Rightarrow \gamma \leq \frac{1}{4LT}$, get $U_5$

\begin{align}
&\frac{1}{T} \sum_{t=1}^T \mathbb{E} ||\theta_{r,n,t-1} - \theta_r||^2 \\ 
&\leq 4 \gamma^2 T \sigma^2 + 32 \gamma^2 T^2 \delta^2 
+ 32 \gamma^2 r^2 \mathbb{E} ||\nabla F(\theta_r) \odot m_{r,n,l}||^2 + 4 w^2 \mathbb{E} ||\theta_r||^2 \\
&\leq 4 \gamma^2 T \sigma^2 + 32 \gamma^2 T^2 \sigma^2 + 32 \gamma^2 T^2 \sum_{i \in S_r} \mathbb{E} ||\nabla F(\theta_r)||^2 + 4 \omega^2 \mathbb{E} ||\theta_r||^2.
\end{align}

Plugging $U_5$ into $U_4$, $U_4$ into $U_3$, $U_3$ into $U_1$, we have:
\begin{align}
&\mathbb{E}\langle \nabla F(\theta_r), \theta_{r+1} - \theta_r \rangle \nonumber\\
&\leq - T \gamma \sum_{i \in S_r} \mathbb{E} || \nabla F^i (\theta_r) ||^2 + \frac{T \gamma}{2} \sum_{i \in S_r} \mathbb{E} || \nabla F^i (\theta_r) ||^2 + 32 \gamma^3 T^3 \frac{N}{|\mathcal{M}^*|} L^2 \sum_{i \in S_r} \mathbb{E} || \nabla F^i (\theta_r) ||^2 \nonumber\\
&+ 8 \omega^2 T \gamma \frac{N}{|\mathcal{M}^*|} L^2 \mathbb{E} || \theta_r ||^2 + 4 \gamma^3 T^2 \frac{N}{|\mathcal{M}^*|} L^2 \sigma^2 + 32 \gamma^3 T^3 \frac{N}{|\mathcal{M}^*|} L^2 \delta^2 + T \gamma \frac{N}{|\mathcal{M}^*|} \delta^2.
\end{align}

Bound $U_2$: 

\begin{align}
    	& \quad \frac{L}{2} \mathbb{E} [|| \theta_{r+1} - \theta_r ||^2] = \frac{L}{2} \sum_{S^p \in S_r} \mathbb{E} || \theta_{r+1}^p - \theta_r^p ||^2 + \frac{L}{2} \sum_{S^p \in K-S_r} \mathbb{E} || \theta_{r+1}^p - \theta_r^p ||^2 \nonumber\\ 
	&= \frac{L}{2} \sum_{S^p \in S_r} \mathbb{E} || \theta_{r+1}^p - \theta_r^p ||^2 = \frac{L}{2} \sum_{S^p \in S_r} \mathbb{E} \left| \left| \frac{1}{|\mathcal{M}_r^p|} \sum_{n \in N_r^p} (\theta_{r,n,0} - \theta_{r,n,T})^p \right| \right|^2 \nonumber\\ 
	&= \frac{L}{2} \sum_{S^p \in S_r} \mathbb{E} \left| \left| \frac{1}{|\mathcal{M}_r^p|} \sum_{n \in N_r^p} \left( \theta_{r,n,0} - \left( \theta_{r,n,0} - \sum_{t=1}^{T} \gamma \nabla F_n (\theta_{r,n,t-1}, \xi_{n,t-1}) \odot m_{n,r} \right) \right)^p \right| \right|^2 \nonumber\\ 
	&= \frac{L}{2} \sum_{S^p \in S_r} \mathbb{E} \left| \left| \frac{1}{|\mathcal{M}_r^p|} \sum_{n \in N_r^p} \sum_{t=1}^{T} \gamma \nabla F_n^p (\theta_{r,n,t-1}, \xi_{n,t-1}) \right| \right|^2 \nonumber\\ 
	&\leq \frac{3L}{2} \sum_{S^p \in S_r} \mathbb{E} \left| \left| \frac{1}{|\mathcal{M}_r^p|} \sum_{n \in N_r^p} \sum_{t=1}^{T} \gamma \left( \nabla F_n^p (\theta_{r,n,t-1}, \xi_{n,t-1}) - \nabla F_n^p (\theta_{r,n,t-1}) \right) \right| \right|^2 \nonumber\\
	&+ \frac{3 \gamma^2 L}{2} \sum_{S^p \in S_r} \mathbb{E} \left| \left| \frac{1}{|\mathcal{M}_r^p|} \sum_{n \in N_r^p} \sum_{t=1}^{T} \nabla F_n^p (\theta_{r,n,t-1}) - \nabla F^p (\theta_r) \right| \right|^2  + \frac{3L}{2} \sum_{S^p \in S_r} \mathbb{E} \left| \left| \frac{1}{|\mathcal{M}_r^p|} \sum_{n \in N_r^p} \sum_{t=1}^{T} \gamma \nabla F_n^p (\theta_r) \right| \right|^2 \nonumber\\ 
	&\leq \frac{3}{2} L T \gamma^2 \frac{N}{|\mathcal{M}^*|} \sigma^2 + 3 L \gamma^2 \frac{N}{|\mathcal{M}^*|} T^2 \left( 4 \gamma^2 T \sigma^2 + 32 \gamma^2 T^2 \delta^2 \right)\nonumber\\
	&+ 32 \gamma^2 T^2 \sum_{S^p \in S_r} \mathbb{E} || \nabla F^p (\theta_r) ||^2 + 4 \omega^2 \mathbb{E} || \theta_r ||^2 + 3 L \frac{N}{|\mathcal{M}^*|} \gamma^2 T \sum_{t=1}^{T} \delta^2 + \frac{3}{2} L \gamma^2 T^2 \sum_{S^p \in S_r} \mathbb{E} || \nabla F^p (\theta_r) ||^2 \nonumber\\ 
	&= \frac{3}{2} L T \gamma^2 \frac{N}{|\mathcal{M}^*|} \sigma^2 + 12 L \gamma^2 \frac{N}{|\mathcal{M}^*|} T^3 \sigma^2 + 96 L \gamma^4 \frac{N}{|\mathcal{M}^*|} T^4 \delta^2 + 96 L^3 \gamma^4 T^4 \frac{N}{|\mathcal{M}^*|} \sum_{S^p \in S_r} \mathbb{E} || \nabla F^p (\theta_r) ||^2 \nonumber\\
	&+ 12 L \gamma^2 T^2 \frac{N}{|\mathcal{M}^*|} \omega^2 \mathbb{E} || \theta_r ||^2 + \frac{3}{2} L \gamma^2 T^2 \sum_{S^p \in S_r} \mathbb{E} || \nabla F^p (\theta_r) ||^2 + 3 L \frac{N}{|\mathcal{M}^*|} \gamma^2 T^2 \delta^2.
\end{align}

Last, we have

\begin{align}
&\mathbb{E}[F(\theta_{R+1})] - \mathbb{E}[F(\theta_1)]\nonumber\\  
&= \sum_{r=1}^R \mathbb{E}[F(\theta_{R+1})] - \sum_{r=1}^R \mathbb{E}[F(\theta_r)] \\
&\leq \sum_{r=1}^R \mathbb{E} \left[ \langle \nabla F(\theta_r), \theta_{R+1} - \theta_r \rangle \right] + \sum_{r=1}^R \frac{L}{2} \mathbb{E} ||\theta_{R+1} - \theta_r||^2.
\end{align}

Plugging $U_1,U_2$ into the above equation, we have:
\begin{align}
&\mathbb{E}[F(\theta_{R+1})] - \mathbb{E}[F(\theta_1)] \nonumber\\
\leq &-T \gamma \sum_{r=1}^R \sum_{i \in S_r} \mathbb{E} ||\nabla F^i(\theta_r)||^2 
+ \frac{T \gamma}{2} \sum_{r=1}^R \sum_{i \in S_r} \mathbb{E} ||\nabla F^i(\theta_r)||^2 + 32 \gamma^3 T^3 \frac{N }{|\mathcal{M}^*|} L^2 \sum_{r=1}^R \sum_{i \in S_r} \mathbb{E} ||\nabla F^i(\theta_r)||^2 \nonumber\\
&+ 8 \omega^2 T \gamma \frac{N }{|\mathcal{M}^*|} L^2 \sum_{r=1}^R \mathbb{E} ||\theta_r||^2 + 4 \gamma^3 T^2 \frac{N }{|\mathcal{M}^*|} L^2 R \sigma^2 + 32 \gamma^3 T^3 \frac{N }{|\mathcal{M}^*|} L^2 R \delta^2 + T \gamma \frac{N }{|\mathcal{M}^*|} L^2 R \delta^2 \nonumber\\
&+ 12 L^3 \gamma^4 T^3 \frac{N }{|\mathcal{M}^*|} \omega^2 \sum_{r=1}^R \mathbb{E} ||\theta_r||^2 + \frac{3}{2} L \gamma^2 T^2 \sum_{r=1}^R \sum_{i \in S_r} \mathbb{E} ||\nabla F^i(\theta_r)||^2 + 3 L \gamma^2 T^2 \frac{N }{|\mathcal{M}^*|} R \delta^2 \\
&\overset{(a)}\leq -\frac{T\gamma}{8} \sum_{r=1}^R \sum_{i \in S_r} \mathbb{E} ||\nabla F^i(\theta_r)||^2 + \left( 8 \omega^2 T \gamma \frac{N }{|\mathcal{M}^*|}  L^2 + 12 L^3 \gamma^2 T^2 \frac{N }{|\mathcal{M}^*|}  \right) \sum_{r=1}^R \mathbb{E} ||\theta_r||^2 \nonumber\\
&+ T \gamma R \frac{N }{|\mathcal{M}^*|}  \left( 32 \gamma^2 T^2 L^2 + 1 + 96 L^3 \gamma^3 T^3 + 3 L \gamma T \right) \delta^2 + \gamma^2 T L R \frac{N }{|\mathcal{M}^*|}  (4 \gamma TL + \frac{3}{2} +12L^2 \gamma^2 t^2) \sigma^2.
\end{align}

where $(a)$ follows because:
\begin{align}
&32 \gamma^2 T^2 r^2 \frac{N_r}{\Gamma_*} L^2 \leq \frac{1}{8} \implies \gamma \leq \frac{\sqrt{\Gamma_*}}{16 T L \sqrt{N}}, \\
&96 L \gamma^3 r^3 T^3 \frac{N_r}{\Gamma_*} \leq \frac{1}{8} \implies \gamma \leq \left( \frac{\Gamma_*}{768 L^3 T N} \right)^{\frac{1}{3}}, \\
&\frac{3}{2} L \gamma T \leq \frac{1}{8} \implies \gamma \leq \frac{1}{12 T L}.
\end{align}

Therefore, we have:
\begin{align}
&\frac{T \gamma}{8} \sum_{r=1}^R \sum_{i \in S_r} \mathbb{E} ||\nabla F^i(\theta_r)||^2 \nonumber\\
&\leq \mathbb{E}[F(\theta_1)] - \mathbb{E}[F(\theta_{R+1})] + \left( 8 \omega^2 T \gamma \frac{N }{|\mathcal{M}^*|}  L^2 + 12 L^3 \gamma^2 T^2 \omega^2\frac{N }{|\mathcal{M}^*|}  \right) \sum_{r=1}^R \mathbb{E} ||\theta_r||^2 \nonumber\\
&+ T \gamma R \frac{N }{|\mathcal{M}^*|}  \left( 32 \gamma^2 T^2 L^2 + 1 + 96 L^3 \gamma^3 T^3 + 3 L \gamma T \right) \delta^2 + \gamma^2 T L R \frac{N }{|\mathcal{M}^*|}  (4 \gamma TL + \frac{3}{2} +12L^2 \gamma^2 T^2) \sigma^2. 
\end{align}

Dividing both sides by $\frac{RT \gamma}{8}$,
\begin{align}
&\frac{1}{R} \sum_{r=1}^R \sum_{i \in S_r} \mathbb{E} ||\nabla F^i(\theta_r)||^2 \nonumber\\
\leq & \frac{8 }{ RT \gamma }\left(\mathbb{E}[F(\theta_1)] - \mathbb{E}[F(\theta_{R+1})] \right) + \left( 64 \omega^2 \frac{N }{|\mathcal{M}^*|}  L^2 + 96 L^3 \gamma T \frac{N }{|\mathcal{M}^*|}  \right) \frac{1}{R}\sum_{r=1}^R \mathbb{E} ||\theta_r||^2 \nonumber\\ +
&\frac{8N}{|\mathcal{M}^*|}  \left( 32 \gamma^2 T^2 L^2 + 1 + 96 L^3 \gamma^3 T^3 + 3 L \gamma T \right) \delta^2 + \gamma L \frac{8N }{|\mathcal{M}^*|}  (4 \gamma TL + \frac{3}{2} +12L^2 \gamma^2 T^2) \sigma^2. 
\end{align}

Until now we complete the proof of Theorem 1.

\end{document}